\documentclass{article}
\usepackage{xcolor,colortbl}
\usepackage{amssymb}
\usepackage{amsmath}
\usepackage{bm}
\usepackage{svg}
\usepackage{adjustbox}

\usepackage{PRIMEarxiv}

\usepackage[utf8]{inputenc} 
\usepackage[T1]{fontenc}    
\usepackage{hyperref}       
\usepackage{url}            
\usepackage{booktabs}       
\usepackage{amsfonts}       
\usepackage{nicefrac}       
\usepackage{microtype}      
\usepackage{lipsum}
\usepackage{fancyhdr}       
\usepackage{graphicx}       
\graphicspath{{media/}}     
\usepackage{verbatim}
\title{MixFace: Improving Face Verification Focusing on Fine-grained Conditions
}

\author{
  Junuk Jung, Sungbin Son, Joochan Park, Yongjun Park, Seonhoon Lee, Heung-Seon Oh\thanks{corresponding author} \\
  School of Computer Science and Engineering\\
  Korea University of Technology and Education (KOREATECH) \\
  
  \texttt{\{rnans33, sbson0621, green669, qkr2938, karma1002, ohhs\}@koreatech.ac.kr} \\
  
}

\begin{document}
\maketitle

\begin{abstract}
	The performance of face recognition (FR) has become saturated for public benchmark datasets such as LFW, CFP-FP, and AgeDB, owing to the rapid advances in CNNs. However, the effects of faces with various fine-grained conditions on FR models have not been investigated because of the absence of such datasets. This paper analyzes their effects in terms of different conditions and loss functions using K-FACE, a recently introduced FR dataset with fine-grained conditions. We propose a novel loss function, MixFace\footnote{Codes are available at https://github.com/Jung-Jun-Uk/mixface.git}, that combines classification and metric losses. The superiority of MixFace in terms of effectiveness and robustness is demonstrated experimentally on various benchmark datasets.
\end{abstract}


\section{Introduction}
Face recognition (FR) is one of the most promising computer vision tasks and is widely utilized in surveillance and security. Recently, FR has achieved remarkable results on benchmark datasets owing to advances in CNNs \cite{resNet, hu2018squeeze, vgg, goingDeeper}. Based on CNNs, there are three major FR research directions: 1) denoising FR data \cite{wang2018devil}, 2) improving face detection and face alignment \cite{deng2020retinaface}, and 3) devising novel loss functions \cite{cao2018vggface2, chopra2005learning, hadsell2006dimensionality, deepface-recognition, schroff2015facenet, taigman2014deepface}. Among them, the third direction has gained the attention of many researchers.

Loss functions in FR can be categorized into two approaches: classification and metric losses. The former optimizes the similarity between the deep feature and weight vectors using class labels. Softmax loss \cite{cao2018vggface2, deepface-recognition, taigman2014deepface} is a typical example of a classification loss. The latter optimizes the similarity between deep feature vectors using pairwise labels. Contrastive loss \cite{chopra2005learning, hadsell2006dimensionality} and triplet loss \cite{schroff2015facenet} belong to the metric loss category. Although both approaches have contributed to improving FR considerably, each has limitations. Softmax loss performs well in the closed-set protocol where test classes are represented in the training set but poorly in the open-set protocol where test classes are disjoint from the training set because of a lack of discriminative power \cite{wen2016discriminative}. Contrastive loss and triplet loss have slow convergence and reach local optima \cite{sohn2016improved} because a limited number of face combinations are used in each training iteration, even over a tremendous number of possible face combinations from a FR dataset. In addition, they cannot distinguish difficult samples because FR models are trained with easy samples in repetition \cite{wang2019multi}.

To address these problems, for classification loss, softmax loss functions using angular margins \cite{deng2019arcface, wang2018cosface, liu2017sphereface, wang2018additive} were presented to obtain a discriminative power satisfying intra-class compactness and inter-class dispersion. In particular, ArcFace \cite{deng2019arcface} improved FR by directly giving margins to the arc degree between the deep feature and target weight vectors. For metric loss, various sampling strategies have been studied to increase the number of face combinations in a training iteration \cite{sohn2016improved} and alleviate the domination of easy samples \cite{wang2019multi, schroff2015facenet, harwood2017smart, ge2018deep}. Both approaches use normalization to the deep feature vectors. Moreover, for classification loss, weight vectors are normalized to obtain a precise cosine similarity space \cite{deng2019arcface, wang2018cosface, wang2018additive, wang2019multi, zhang2019adacos, yu2019deep}, and a scale factor  is used to distinguish subtle differences between these similarities. Research on classification loss has gained more attention because it is difficult to alleviate inefficiency in metric loss. Accuracy greater than 99\% accuracy has been achieved on several benchmark datasets (LFW \cite{huang2008labeled}, CFP-FP \cite{sengupta2016frontal}, and AgeDB \cite{moschoglou2017agedb}).

However, it has been infeasible to investigate the effects of faces with fine-grained conditions such as varying lighting, poses, accessories, and facial expressions on FR in previous research due to the lack of such datasets. An in-depth analysis of fine-grained conditions is essential to improve robust FR models. Recently, the K-FACE \cite{choi2021k} dataset, which contains fine-grained conditions, was introduced to improve FR research. We analyzed the performance of classification and metric loss functions on this dataset. The analysis showed that the two approaches lack the representation power to adequately accommodate the changes in conditions, but we hypothesized that the two approaches could be combined. 

Therefore, we propose a novel loss function, MixFace, that combines classification and metric losses. MixFace uses scale factors $s_1$ and $s_2$ for the two approaches. We assumed that manual tuning would not guarantee proper harmony in MixFace because they have different ranges. Therefore, we present a unified scale factor $\epsilon$ derived from ideal similarities of the classification and metric losses such that a hyper-parameter easily determines $s_1$ and $s_2$.  Our experimental results demonstrate the superiority of MixFace in terms of two aspects. First, MixFace is effective because it outperforms the existing loss functions with fine-grained conditions. Second, MixFace maintains generalization capability because it achieves competitive performance on benchmark datasets (LFW, CFP-FP, and AgeDB30). The contributions of this paper can be summarized as follows:
\begin{itemize}
  \item This paper examines the effects of fine-grained conditions in FR using K-FACE in terms of both training and testing, as well as classification and metric losses.
  \item This paper proposes a novel loss function, MixFace, that combines classification and metric losses. It inherits the benefits and complements the weaknesses of the two approaches. Exhaustive experiments show the superiority of MixFace in terms of effectiveness and robustness.
\end{itemize}

\begin{figure}[!htbp]
    \centering
   \includegraphics[width=13cm]{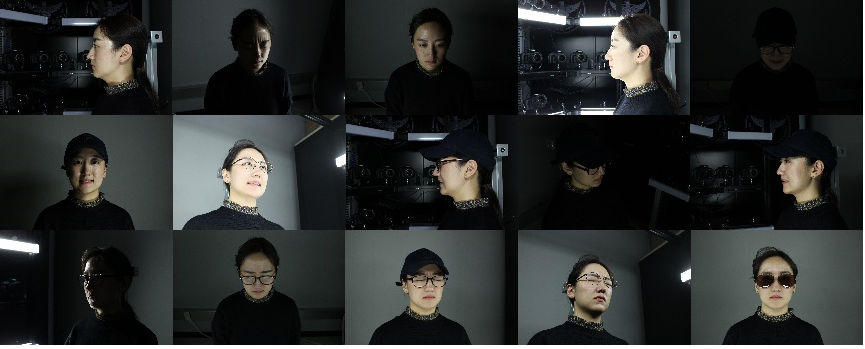}
   \hfil
\caption{K-FACE sample images with various conditions.}
\label{fig:kface_sample}
\end{figure}

\section{K-FACE Dataset}
\subsection{Summary of the K-FACE}

K-FACE \cite{choi2021k} is a FR dataset consisting of 17,550,000 $(1000\times6^+\times29\times3\times20)$ images of 1,000 subjects with more than six accessories, 35 lights , three facial expressions, and 27 poses. Owing to privacy and disclosure issues, only a subset of K-FACE consisting of 4,176,600 $(400\times6\times29\times3\times20)$ images of 400 subjects with six accessories, 29 lights, three facial expressions, and 20 poses are available. A condition indicates an instance of four attributes and has a fine granularity due to many combinations. Table \ref{tab:config}  summarizes the configuration of the dataset we used, and sample images with various conditions are shown in Figure \ref{fig:kface_sample}. In the remainder of this paper, “K-FACE” denotes the subset used. 

\begin{table}[!htbp]
    \caption{Configuration of K-FACE.}
    \centering
    \includegraphics[width=13cm]{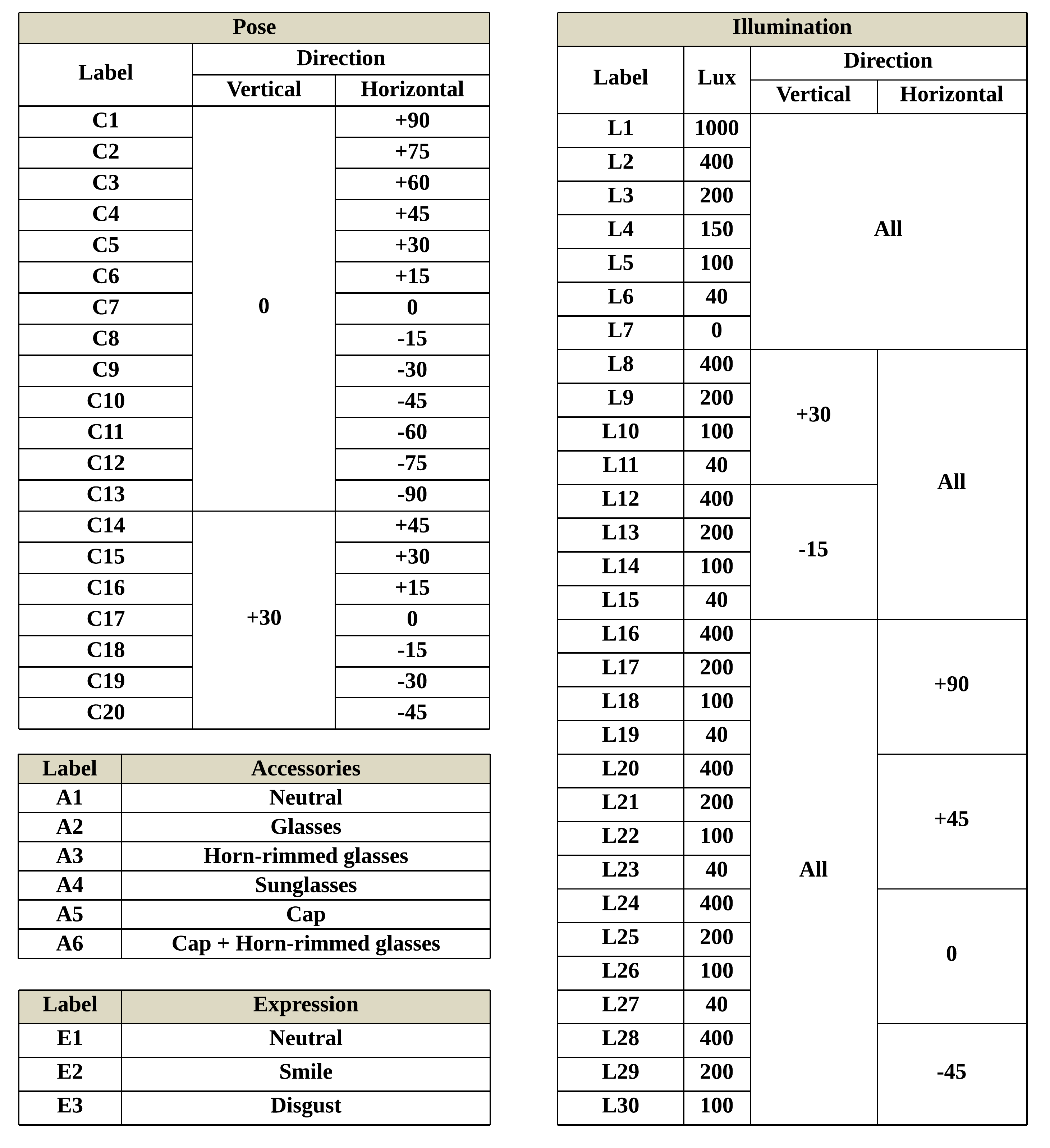}
    \label{tab:config}
\end{table}

\begin{table}[!htbp]
    \caption{Training and test datasets with different variances of conditions. \textbf{A}ccessories, \textbf{L}ux, \textbf{E}xpression, and
    \textbf{P}ose.}
    \label{tab:training and test datasets}
      \centering
        \begin{tabular}[width=12.5cm]{| c | c | c | c | c | c | c |}
         \rowcolor{gray!50}
         \hline
         \textbf{Train ID} & \textbf{A} & \textbf{L} & \textbf{E} & \textbf{P} & \textbf{\#Images} & \textbf{Variance} \\
         \hline
         $\text{T}_1$ & $\text{A}1$ & $1000$ & $\text{E}1$ & $\text{C}4-10$ & 2,590 & Very Low\\
         
         $\text{T}_2$ & $\text{A}1-2$ & $400-1000$ & $\text{E}1$ & $\text{C}4-10$  & 46,620 & Low\\ 
         
         $\text{T}_3$ & $\text{A}1-4$ & $200-1000$ & $\text{E}1-2$ & $\text{C}4-13$  & 654,160 & Middle\\ 
         
         $\text{T}_4$ & $\text{A}1-6$ & $40-1000$ & $\text{E}1-3$ & $\text{C}1-20$  & 3,862,800 & High\\ 
         \hline
         \rowcolor{gray!50}
         \textbf{Test ID} & \textbf{A} & \textbf{L} & \textbf{E} & \textbf{P} & \textbf{\# Pairs} & \textbf{Variance} \\
         \hline
         $\text{Q}_1$ & $\text{A}1$ & $1000$ & $\text{E}1$ & $\text{C}4-10$ & 1,000 & Very Low\\ 
         
         $\text{Q}_2$ & $\text{A}1-2$ & $400-1000$ & $\text{E}1$ & $\text{C}4-10$  & 100,000 & Low\\ 
         
         $\text{Q}_3$ & $\text{A}1-4$ & $200-1000$ & $\text{E}1-2$ & $\text{C}4-13$  & 100,000 & Middle\\ 
         
         $\text{Q}_4$ & $\text{A}1-6$ & $40-1000$ & $\text{E}1-3$ & $\text{C}1-20$  & 100,000 & High\\ 
         \hline
        \end{tabular}
\end{table}

\subsection{Data Preparation}

Of the 400 subjects in the dataset , 370 and 30 subjects were training and test data, respectively. To analyze the effects of the variances of conditions, four datasets of training and test sets with different degrees of condition variation were constructed, as shown in Table \ref{tab:training and test datasets}. $\text{T}_1-\text{T}_4$ and $\text{Q}_1-\text{Q}_4$ denote  training and test datasets, where the variance of conditions increases with increasing subscript index. The variance of the conditions increases as more conditions (values) are included. Each test dataset consists of positive and negative pairs with a 1:1 ratio, where the positive indicates the two face images belong to the same class, and the negative indicates that they do not. They were randomly sampled without replacement using four attributes from 30 subjects. Table 2 summarizes the training and test datasets with different variances of conditions.

\section{Observations}
This section analyzes the effects of variances of conditions on the performance of two different loss functions through concrete experiments for face verification, where performance was measured by accuracy. ArcFace \cite{deng2019arcface} and SN-pair loss were chosen for baseline classification and metric losses, respectively, where the SN-pair loss is a similarity-based N-pair loss \cite{sohn2016improved}. ResNet-34 \cite{resNet} was chosen as the baseline backbone. The details of the two loss functions and implementation are explained in Sections 4 and 5, respectively.
\subsection{Effects of  Variances of Conditions}
\begin{figure}[!htbp]
\centering
   \includegraphics[width=13cm]{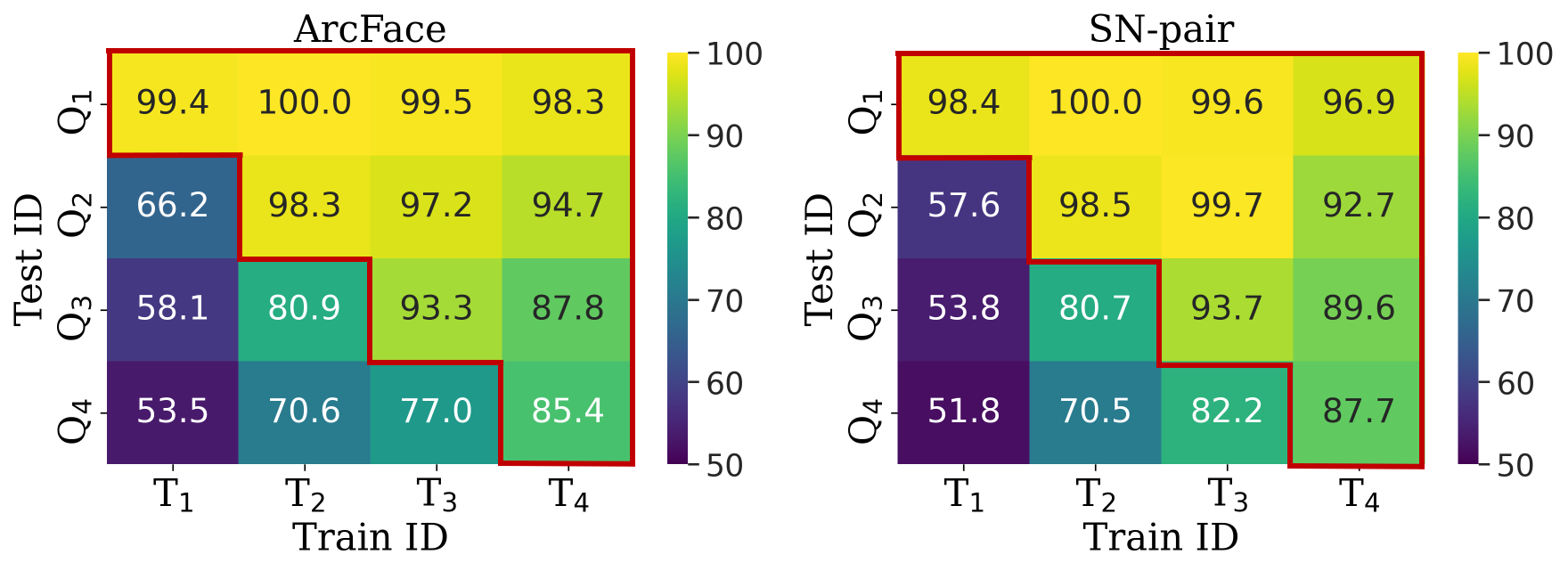}
   \hfil
\caption{Heatmaps by varying training and test datasets in ArcFace and SN-pair.}
\label{fig:various_conditions}
\end{figure}

Figure \ref{fig:various_conditions} shows the results obtained with  $\text{T}_1-\text{T}_4$ and $\text{Q}_1-\text{Q}_4$ as heatmaps; for example, $(\text{T}_1,\text{Q}_2,66.2)$ in ArcFace means that the accuracy is 66.2 on $\text{Q}_2$ with a FR model trained with $\text{T}_1$. Analysis was conducted by partitioning the heatmaps into a lower triangle $\text{P}_\text{L}$ (no border) and an upper triangle $\text{P}_\text{U}$ (red border). Simply, $\text{P}_\text{L}$ and $\text{P}_\text{U}$ indicate undertraining and overtraining, respectively, by specifying with and without conditions. Undertraining indicates that some conditions are excluded in the training datasets but included in the test datasets; vice versa for overtraining. We observed several key results. First, undertraining must be avoided because of its low performance. Second, the difficulty increases as the variance of conditions increases in the test datasets. In addition, the training dataset with a high variance of conditions, $\text{T}_4$, results in underperformance in all test datasets comparted with others such as $\text{T}_2$ and $\text{T}_3$. This shows that overtrained models lack representation power from all variances of conditions. In summary, for a robust FR with fine-grained conditions, the representation power of the model must be increased to minimize the performance penalty that occurs on datasets from low to high variance of conditions. 

\subsection{Classification vs. Metric Losses}
\begin{figure}[!htbp]
\centering
   \includegraphics[width=13cm]{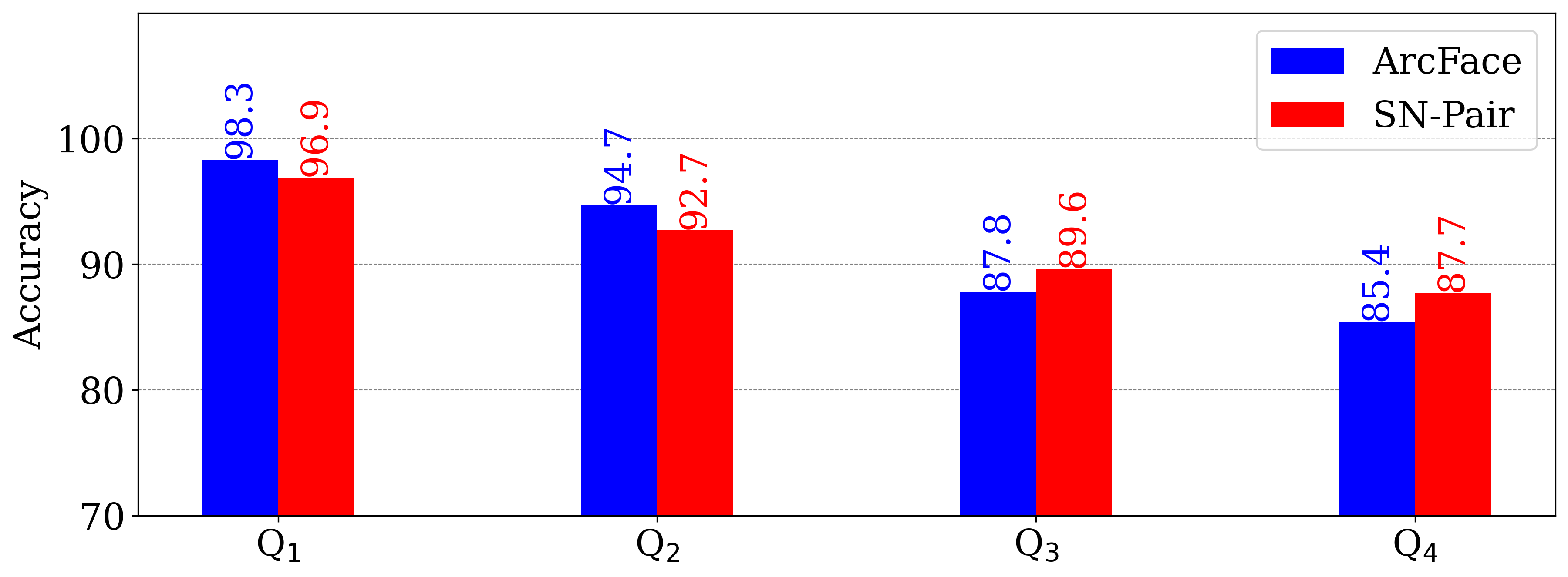}
   \hfil
\caption{Performance of classification and metric losses on different test datasets.}
\label{fig:class vs metric}
\end{figure}

Figure \ref{fig:class vs metric} shows the performance of the classification and metric losses. The test datasets in Table \ref{tab:training and test datasets} were evaluated using models trained with $\text{T}_4$. Interestingly, ArcFace shows improvements over SN-Pair in $\text{Q}_1$ and $\text{Q}_2$, whereas SN-pair achieves better performance in $\text{Q}_3$ and $\text{Q}_4$. This indicates that classification loss has strength on low variances, whereas metric loss has strength on high variances. This reveals the importance of metric loss even though research on classification loss became dominant because its superiority was demonstrated on benchmark datasets. This observation motivated the invention of MixFace.

\section{Proposed Loss Function}
In this section, after introducing representative loss functions for classification and metric losses, the proposed loss function inheriting their advantages is described.

\subsection{Classification Loss: ArcFace}
$\bm{x}_i$ and $y_i$ are a deep feature vector and class-label of $i$-th face image, respectively, and $\bm{w}_j$ is a class weight vector of $j$-th class. Then, softmax loss is defined as:

\begin{equation}
    \mathcal{L}_{softmax} = -\frac{1}{N}\sum_{i=1}^{N}\log{\frac{e^{\bm{w}^\intercal_{y_i}\bm{x}_i}}{\sum_{j=1}^{C} e^{\bm{w}^\intercal_j\bm{x}_i}}}
\end{equation}
where $C$ is the number of classes and $N$ is the size of the batch. 

\noindent The research on center loss \cite{wen2016discriminative} has indicated that softmax loss is unsuitable for FR because it does not sufficiently satisfy intra-class compactness and inter-class dispersion owing to the lack of discriminative power. Thus, the recently developed classification losses \cite{deng2019arcface, wang2018cosface, liu2017sphereface, wang2018additive} have attempted to increase the discriminative power by incorporating angle margins into $\cos{\theta_{i,y_i}}=\bm{w}^\intercal_{y_i}\bm{x}_i/(\lVert \bm{w}^\intercal_{y_i} \rVert \lVert \bm{x}_i \rVert)$ in various ways. Note that $\cos{\theta_{i,y_i}}$ is the angle between the deep feature and target weight vectors. ArcFace \cite{deng2019arcface} is a representative classification loss based on softmax using an angular margin:  

\begin{equation}
    \mathcal{L}_{arc} = -\frac{1}{N}\sum_{i=1}^{N}\log{\frac{e^{s_1\cdot\cos{(\theta_{i,y_i} + m)}}}{e^{s_1\cdot\cos{(\theta_{i,y_i} + m)}}+\sum_{j=1, j\neq y_i}^{C}e^{s_1\cdot\cos{\theta_{i,j}}}}}
\label{eq:arcface}
\end{equation}
where $s_1$ is a scale factor and $m$ is an angular margin.

\subsection{Metric Loss: SN-Pair loss}
When using metric loss, our concern is $\cos{\theta_{i,j}}=\bm{x}^\intercal_{j}\bm{x}_i/(\lVert \bm{x}^\intercal_{j} \rVert \lVert \bm{x}_i \rVert)\in \mathbb{R}^{N\times N}$ where $\cos{\theta_{i,j}}$ is the angle between the deep feature vectors of the $i$-th and $j$-th face images. A pair of two face images is a positive example if they indicate the same person, and a negative example otherwise. They can be extracted from the upper triangle matrix of $\cos{\theta_{i,j}}$ as:

\begin{equation}
    \begin{aligned}
    \{\cos{\theta^{p}_{k}}\}=\{\cos{\theta_{i,j}} \ \text{for} \ i<j\ \ \text{where} \ y_i = y_j\} \\
    \{\cos{\theta}^{n}_{l}\}=\{\cos{\theta_{i,j}} \ \text{for} \ i<j\ \ \text{where} \ y_i \neq y_j\} 
    \end{aligned}
\end{equation}
where $p$ and $n$ are positive and negative pairs, respectively, and $k$ and $l$ are the indices of the positive and negative pairs, respectively. 

\noindent N-pair loss \cite{sohn2016improved} was devised to compute the loss of samples in a batch at once to alleviate the problem of insufficient pairs at each training iteration in triplet loss \cite{schroff2015facenet}. Recent metric loss functions \cite{sun2020circle, wang2019multi, yu2019deep} have employed cosine similarity to reduce the metric gap between training and testing. As a result, the similarity-based N pair loss (SN-pair loss) is defined as:

\begin{align}
\mathcal{L}_{sn-pair}=\frac{1}{K}\sum^{K}_{k=1}\log{(1+\sum^{L}_{l=1}e^{s_2\cdot\cos{\theta}^{n}_{l}-s_2\cdot\cos{\theta^{p}_{k}}})} =  -\frac{1}{K}\sum_{k=1}^{K}\log{\frac{e^{s_2\cdot\cos{\theta}^{p}_{k}}}{e^{s_2\cdot\cos{\theta}^{p}_{k}}+\sum_{l=1}^{L} e^{ s_2\cdot\cos{\theta}^{n}_{l}}}}
\label{eq:sn-pair}
\end{align}
where $K$ and $L$ are the numbers of positive and negative examples, respectively, and $s_2$ is a scale factor. 

\subsection{MixFace}
We have observed that the classification loss has strength on the low variances of conditions, whereas the metric loss has it on the high variances (Section 3). We propose that combining the two losses inherits the benefits and complements each other. As a result, we propose MixFace:

\begin{equation}
    \mathcal{L}_{mix} = \mathcal{L}_{arc} + \mathcal{L}_{sn-pair}
\end{equation}
There are two scale factors, $s_1$ and $s_2$, applicable to ArcFace and SN-Pair loss, respectively. It is essential to adjust the scale factors to improve the performance. Optimizing them using grid and random searches\cite{bergstra2012random} is problematic because it requires huge computational costs owing to their different behaviors. To solve this problem, we devised a unified scale factor used in both ArcFace and SN-Pair loss without loss of generality. The ideal similarities of the positives and negatives are expected to be $\cos{(\theta_{i,y_i}+ m)}\rightarrow 1$ and $\cos{\theta_{i,j \neq y_i}}\rightarrow 0$ in ArcFace and, similarly, $\cos{\theta}^{p}_{k} \rightarrow 1$ and $\cos{\theta}^{n}_{l} \rightarrow 0$ in SN-Pair loss. The cosine values in the logit of Equations \ref{eq:arcface} and \ref{eq:sn-pair} are replaced with the ideal values as the probability of a target class in ArcFace and the SN-pair becomes 1. 

\begin{equation}
\begin{aligned}
1 \approx \frac{e^{s_1\cdot\cos{m}}}{e^{s_1\cdot\cos{m}} + C-1} = 1-\epsilon \\ 1 \approx \frac{e^{s_2}}{e^{s_2} + L} = 1-\epsilon
\end{aligned}
\end{equation}
where $\epsilon$ is a hyperparameter with a small value to avoid an infinite problem.

\noindent By rearranging the equations above, the scale factors $s_1$ and $s_2$ are derived with $\epsilon$:
\begin{gather}
s_1 = \frac{\log{(1-\epsilon)} + \log{(C-1)} -\log{\epsilon}}{\cos{m}}, \ s_2 = \log{(1-\epsilon)} + \log{L} -\log{\epsilon}
\end{gather}
Here we define $\epsilon$ as a unified scale factor.

\section{Experiments}

\subsection{Implementation Details}
\textbf{Preprocessing.} MS1M-RetinaFace (MS1M-R) \cite{deng2020retinaface} and K-FACE \cite{choi2021k} datasets were used. MS1M-R is a refinement of MS1M \cite{guo2016ms} with 5.1M images and 93K identities. The K-FACE dataset that we used is summarized in Table \ref{tab:training and test datasets}. Face regions and landmarks were extracted from all the images in MS1M-R and K-FACE using RetinaFace \cite{deng2020retinaface}. Then, face alignment was performed based on similarity transformation. All the images were resized to $112 \times 112$ and normalized using means $[0.485,0.456,0.406]$ and standard deviations $[0.229,0.224,0.225]$. In K-FACE, there are errors in face detection owing to high darkness and low illumination. Fortunately, the face images have similar bounding boxes because they were taken in a strictly controlled environment. Therefore, the bounding boxes and landmarks of erroneous face images were replaced with those of the bright images in L1.  

\noindent\textbf{Training.} ResNet-34 \cite{resNet} was chosen as the backbone model. The mini-batch size was set to 512. For the metric loss, 256 positive pairs were randomly sampled from MS1M-R to avoid domination of negative pairs due to a large number of classes, whereas 512 images were randomly sampled from K-FACE because of the small number of classes. The SGD optimizer was employed with a momentum of 0.9 and a weight decay of 0.0005. A cosine annealing scheduler was utilized with a maximum epoch of 20, a warm-up epoch of 3, and an initial learning rate of 0.1.

\noindent\textbf{Testing.} The evaluation aimed at face verification, by checking whether two face images belonged to the same class or not. For a face image, a deep feature vector concatenating those of the original and horizontal flip images was used. Two face images were considered to belong to the same class if the cosine similarity between the two images was above a threshold . Performance was measured based on verification accuracy recorded after finding the best threshold in a test dataset automatically. 

\subsection{Evaluation Results}
\noindent\textbf{Model comparison on K-FACE.} Table \ref{tab:comparsion on K-FACE} shows the comparison of MixFace and other loss functions on K-FACE. The models were evaluated using  $\text{Q}_1-\text{Q}_4$ after training with $\text{T}_4$and selecting the best-performing models on $\text{Q}_4$ as a baseline. FixCos\cite{zhang2019adacos} and AdaCos\cite{zhang2019adacos} are a hyper-parameter free classification loss. The hyper-parameters for ArcFace were set to $s_1=16,m=0.25$ according to our heuristics. The scale factors in Norm-softmax, CosFace, and SN-pair loss were set to 16. The margin in CosFace was set to 0.25. In the multi-similarity loss \cite{wang2019multi}, MS-loss, the hyper-parameters $\alpha$, $\gamma$, and $\beta$ were set to 2, 0.5, and 50, respectively. MixFace achieved the best performance on all test datasets, ranging from low to high variances of conditions.

\begin{table}[!htb]
    \caption{Model comparison on K-FACE}
    \label{tab:comparsion on K-FACE}
      \centering
        \begin{tabular}[width=12.5cm]{| c | c  c  c  c |}
         \hline
         Loss function & $\text{Q}_1$ & $\text{Q}_2$ &  $\text{Q}_3$ & $\text{Q}_4$\\
         \hline
         FixCos & 98.11 & 90.47 & 84.88 & 80.81 \\
         ArcFace & 98.30 & 94.77 & 87.87 & 85.41\\ 
         CosFace & 98.20 & 94.00 & 89.76 & 86.57\\ 
         AdaCos & 99.70 & 95.97 & 91.17 & 87.69 \\
         
         \hline
         N-pair loss & 92.40 & 90.06 & 88.95 & 87.15\\
         SN-pair loss& 96.90 & 92.69 & 89.58 & 87.70\\
         Ms-loss & 99.10 & 96.30 & 91.43 & 88.82\\
         SN-pair loss, $(s_2=64)$ & 99.20 & 95.01 & 91.84 & 89.74 \\
         \hline
         MixFace & \textbf{100} & \textbf{96.37} & \textbf{92.36} & \textbf{89.80}\\
         \hline
        \end{tabular}
\end{table}

\noindent\textbf{Unified scale factor.} Table \ref{tab:unfied scale factor} shows the effectiveness of the proposed unified scale factor. For a fair experiment, we set the unified scale factor $\epsilon$ to $1\times10^{-2}$ and $1\times10^{-22}$, similar to the value of scale factor $s_2$ of MixFace,  Their settings  are approximately $(10.84, 16.37)$ and $(58.83, 62.43)$, respectively. The FR model was not properly trained when $s_1$ and $s_2$ were set to (64, 64). The proposed unified scale factor improved the performance in all test datasets compared to using the scale factors $s_1$ and $s_2$ independently. These experimental results demonstrate the validity and superiority of the proposed method for adjusting the scale factor ratio for each loss function.

\begin{table}[!htb]
    \caption{Effects of unified scale factor}
    \label{tab:unfied scale factor}
      \centering
        \begin{tabular}[width=12.5cm]{| c |  c  c  c  c |}
         \hline
         Loss function & $\text{Q}_1$ & $\text{Q}_2$ &  $\text{Q}_3$ & $\text{Q}_4$\\
         \hline
         MixFace $s_1, s_2=(16,16)$ & 96.80 & 93.33 & 89.80 & 87.33\\ 
         MixFace $\epsilon=1\times10^{-2}$ & \textbf{98.40} & \textbf{94.06} & \textbf{90.19} & \textbf{88.12}\\ 
         \hline
         MixFace $s_1, s_2=(16,64)$ & 98.50 & 94.06 & 90.55 & 89.45\\
         MixFace $s_1, s_2=(64,64)$ & \multicolumn{4}{c|}{The model is not properly trained}\\
         MixFace $\epsilon=1\times 10^{-22}$ & \textbf{100} & \textbf{96.37} & \textbf{92.36} & \textbf{89.80}\\
         \hline
        \end{tabular}
\end{table}

\begin{figure}[!htbp]
\centering
   \includegraphics[width=12.5cm]{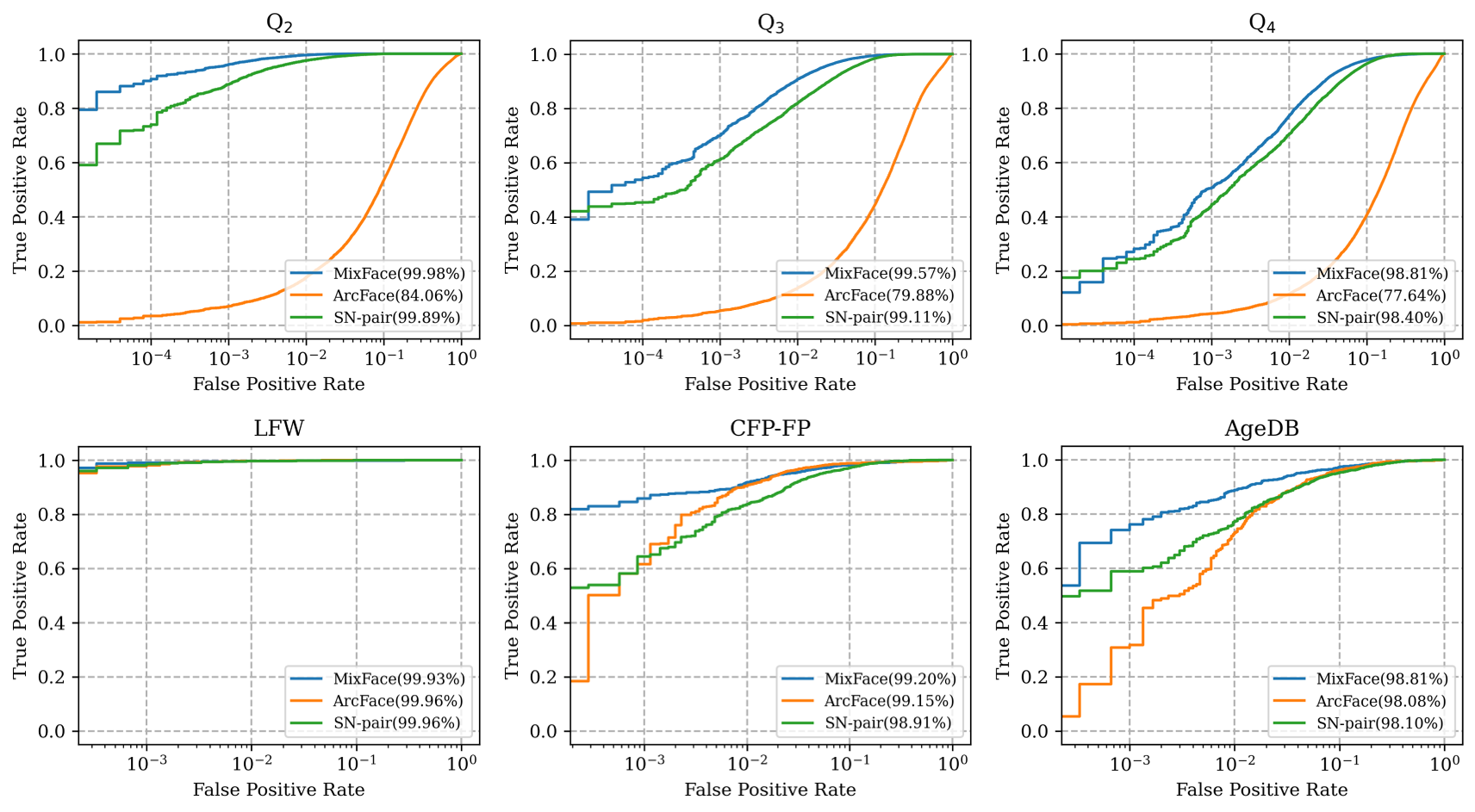}
   \hfil
\caption{Comparison of ROC curves of models trained with MS1M-R+$\text{T}_4$}
\label{fig:roc_curve}
\end{figure}

\begin{figure}[!htbp]
\centering
   \includegraphics[width=13cm]{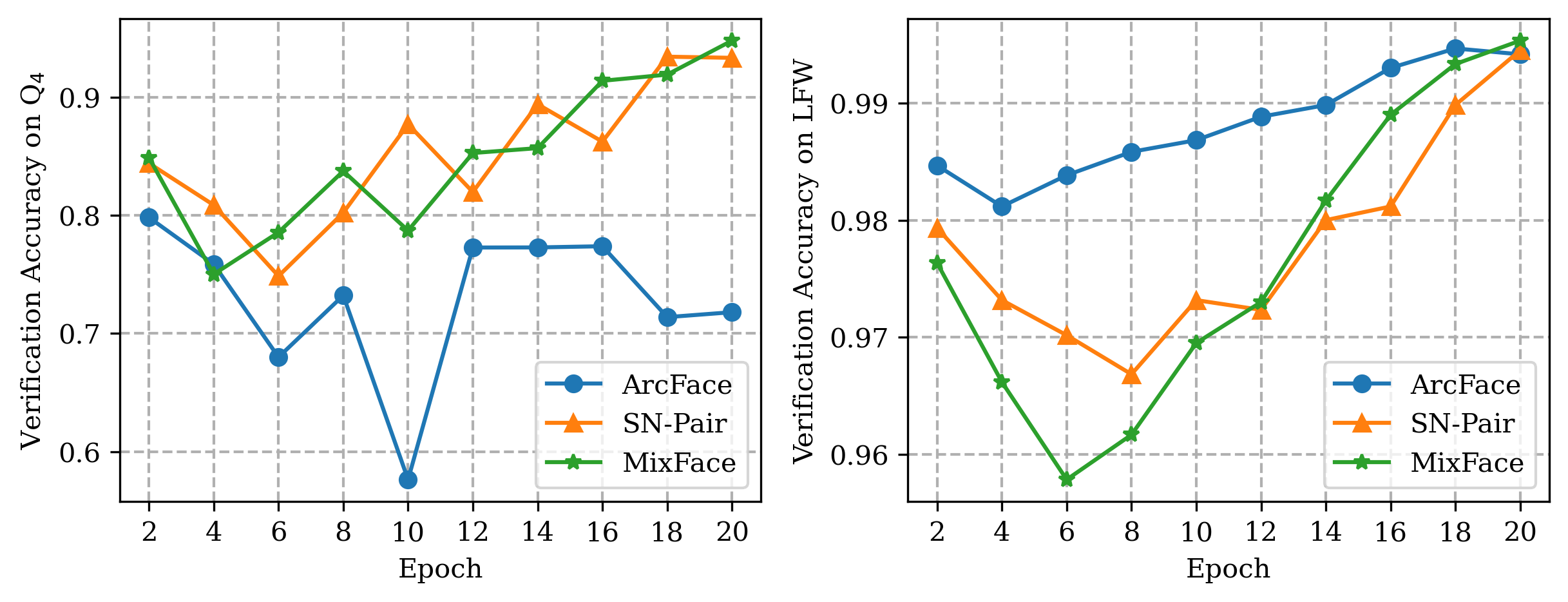}
   \hfil
\caption{Verification accuracies of epochs on models trained with MS1M-R+$\text{T}_4$}
\label{fig:epoch_per_acc}
\end{figure}

\noindent\textbf{Comparison with benchmark datasets.} Table \ref{tab:comparsion bench} shows the performance of the three loss functions using various training and test datasets with the same training protocol. Test datasets were evaluated with the best-performing models on LFW. The first three lines were obtained from models trained using MS1M-R. We can see the performance of ArcFace $>$ MixFace $>$ SN-pair loss for all the test datasets. It shows undertraining because the performance of the three models significantly degraded as the variance of conditions increased owing to the rare existence of extreme conditions in MS1M-R. Interestingly, except for CFP-FP, MixFace $>$ SN-pair $>$ ArcFace according to the second three lines trained with MS1M-R+T4, including fine-grained conditions. This shows that MixFace has strength in fine-grained conditions while also being competitive on benchmark datasets in unconstrained environments. Figure \ref{fig:roc_curve} shows the superiority of MixFace using the ROC curves as the ratios between true-positive and false-positive rates are higher than others overall. Figure \ref{fig:epoch_per_acc} describes verification accuracies of epochs. Notably, MixFace and SN-pair show similar behaviors on Q4 and LFW, whereas ArcFace produced a competitive performance on LFW but was poor on Q4. It shows the lack of representation power of ArcFace on high variance conditions.

\begin{table}[!htb]
    \caption{Comparison of three loss functions on different test datasets}
    \label{tab:comparsion bench}
      \centering
        \begin{tabular}[width=12.5cm]{| c | c | c  c  c | c  c  c |}
         \hline
         Loss functions & Train Datasets & $\text{Q}_2$ & $\text{Q}_3$ & $\text{Q}_4$ & LFW & CFP-FP &  AgeDB \\
         \hline
         ArcFace  & MS1M-R & \textcolor{blue}{\textbf{98.71}} & \textcolor{blue}{\textbf{86.60}} & \textcolor{blue}{\textbf{82.03}} & \textcolor{blue}{\textbf{99.80}} & \textcolor{blue}{\textbf{98.41}} & \textcolor{blue}{\textbf{98.08}} \\
         SN-pair loss  & MS1M-R & 92.85 & 76.36 & 70.08 & 99.55 & 96.20 & 95.46 \\
         MixFace  & MS1M-R & 97.36 & 82.89 & 76.95 & 99.68 & 97.74 & 97.25 \\
         \hline
         \hline
         ArcFace  & MS1M-R+$\text{T}_4$ & 76.58 & 73.13 & 71.38 & 99.46 & \textcolor{red}{\textbf{96.75}} & 93.83 \\
         SN-pair loss  & MS1M-R+$\text{T}_4$ & 98.37 & 94.98 & 93.33 & 99.45 & 94.90 & 93.45 \\
         MixFace  & MS1M-R+$\text{T}_4$ & \textcolor{red}{\textbf{99.27}} & \textcolor{red}{\textbf{96.85}} & \textcolor{red}{\textbf{94.79}} & \textcolor{red}{\textbf{99.53}} & 96.32 & \textcolor{red}{\textbf{95.56}} \\
         \hline
        \end{tabular}
\end{table}


\subsection{Effects of Unseen Conditions}
\begin{figure}[!htbp]
\centering
   \includegraphics[width=12.5cm, height=3cm]{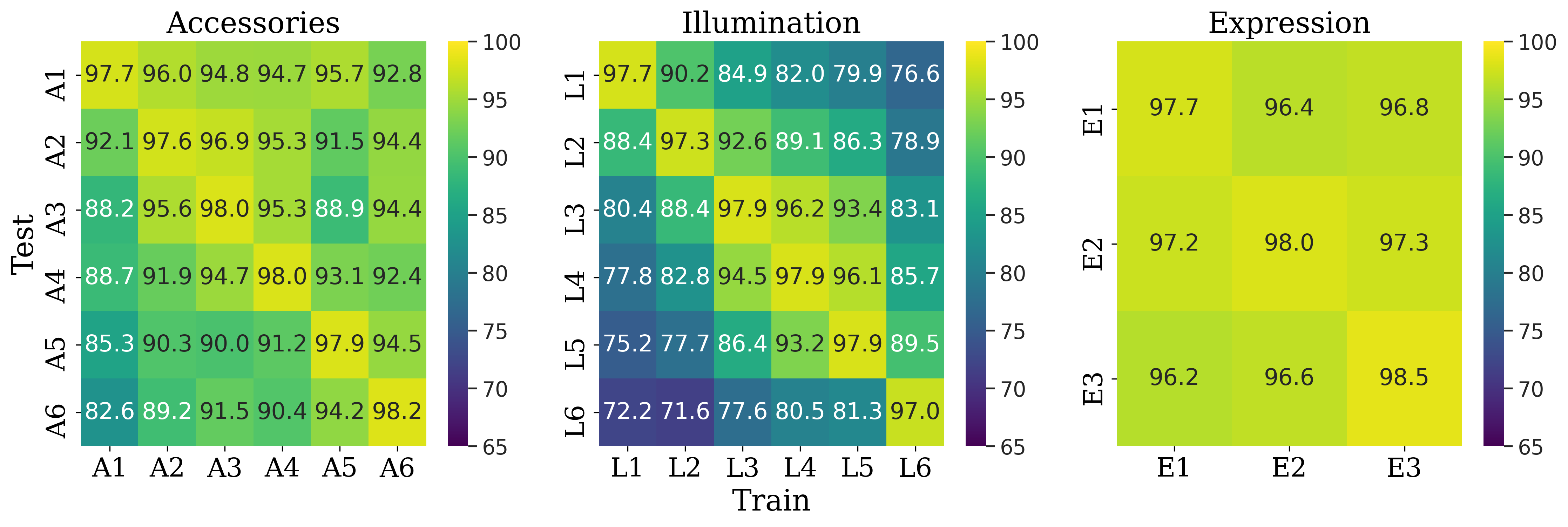}
   \hfil
\caption{Heatmaps of changing single conditions from (A1,L1,E1,C1-20) (see Table \ref{tab:config}). The x- and y-axes denote conversion of a specific condition from the original in training and test datasets, respectively. For the test datasets, we randomly generated 10,000 unique pairs, then converted the corresponding conditions to the y-axis symbols.}
\label{fig:singe_conditions}
\end{figure}

Figure \ref{fig:singe_conditions} shows the heatmaps\footnote{Similar results were observed using ArcFace and SN-pair, respectively.} using MixFace with respect to variation of different conditions. Consider a diagonal line $\text{P}_\text{D}$ and a non-diagonal line $\text{P}_\text{N}$ of a heatmap. In $\text{P}_\text{D}$ of the three heatmaps, the differences between the highest and lowest accuracies were small and in the order of  accessories (+0.6\%) $\approx$ illumination (+0.8\%) $\approx$ expression (0.9\%). This means that the effect of different conditions of a type on the FR is small. However, the differences between the average accuracies of $\text{P}_\text{D}$ and $\text{P}_\text{N}$ are relatively large in the order of illumination (+13.1\%) $>$ accessories (+5.6\%) $>$ expression (+1.3\%). This implies that including unseen conditions in both training and testing considerably affects FR. 


\section{Conclusion}
This paper proposes MixFace, which inherits the benefits of classification and metric losses by analyzing their performance with respect to the variances of conditions in the K-FACE dataset. The superiority of MixFace was demonstrated by a series of experiments on K-FACE and other benchmark datasets, showing that it has strength on various degrees of variance of conditions while maintaining competitiveness on benchmark datasets. In MixFace, manual tuning of hyperparameters can be avoided by presenting a unified scale factor. We expect MixFace to be a good option for surveillance tasks with fine-grained conditions. In our future work, we plan to investigate various methods of mixing classification and metric losses. 

\section{Acknowledgement}
This work was supported by the National Research Foundation of Korea (NRF) grant funded by the Korea government (MSIT) (No. NRF-2019R1G1A1003312) and (No. NRF-2021R1I1A3052815).

\bibliographystyle{unsrt}  
\bibliography{mixface_prime_arxiv}

\end{document}